\documentclass[lettersize,journal]{IEEEtran}
\usepackage{amsmath,amsfonts}
\usepackage{algorithm}
\usepackage{algpseudocode}
\usepackage{array}
\usepackage[caption=false,font=footnotesize,labelfont=rm,textfont=rm]{subfig}
\usepackage{textcomp}
\usepackage{stfloats}
\usepackage{url}
\usepackage{verbatim}
\usepackage{graphicx}
\usepackage{tikz}
\usepackage{amssymb}
\usepackage{booktabs}
\usepackage{multirow}
\usepackage{comment}
\usepackage{xcolor}
\newtheorem{remark}{Remark}

\usepackage{soul}
\usepackage{hyperref}
\usepackage{bbm}
\usepackage[framemethod=tikz]{mdframed}
\usepackage[normalem]{ulem}
\makeatletter
\def\endthebibliography{%
	\def\@noitemerr{\@latex@warning{Empty `the bibliography' environment}}%
	\endlist
}
\makeatother

\begin{document}

\title{Integrating Decision-Making Into Differentiable Optimization Guided Learning for End-to-End Planning of Autonomous Vehicles }

\author{
    Wenru Liu, Yongkang Song, Chengzhen Meng, Zhiyu Huang, Haochen Liu, Chen Lv, and Jun Ma
    \thanks{Wenru Liu, Chengzhen Meng, and Jun Ma are with the Robotics and Autonomous Systems Thrust, The Hong Kong University of Science and Technology, China (e-mail: cmeng403@connect.hkust-gz.edu.cn; wliu354@connect.hkust-gz.edu.cn; jun.ma@ust.hk).} 
         \thanks{Yongkang Song is with Lotus Robotics, China (e-mail: yongkang.song@lotuscars.com.cn).} 
    \thanks{Zhiyu Huang, Haochen Liu, and Chen Lv are with the School of Mechanical and Aerospace Engineering, Nanyang Technological University, Singapore 639798 (e-mail: zhiyu001@e.ntu.edu.sg; haochen002@e.ntu.edu.sg; lyuchen@ntu.edu.sg).} 
	}



\maketitle

\begin{abstract}  
Expert demonstrations are crucial for learning-based autonomous driving (AD) systems, but the performance could be limited if the autonomous vehicles (AVs) merely replicate those demonstrations. 
To achieve optimized driving performance for high-level AD, effective decision-making and trajectory planning are essential for AVs to generate motion plans that enhance driving performance beyond imitation from expert demonstrations.
We address this decision-making capability within an end-to-end planning framework that focuses on motion prediction, decision-making, and trajectory planning. 
Specifically, we formulate decision-making and trajectory planning as a differentiable nonlinear optimization problem, which ensures compatibility with learning-based modules to establish an end-to-end trainable architecture. 
This optimization introduces explicit objectives related to safety, traveling efficiency, and riding comfort, guiding the learning process in our proposed pipeline.
Intrinsic constraints resulting from the decision-making task are integrated into the optimization formulation and preserved throughout the learning process.
By integrating the differentiable optimizer with a neural network predictor, the proposed framework is end-to-end trainable, aligning various driving tasks with ultimate performance goals defined by the optimization objectives.
The proposed framework is trained and validated using the Waymo Open Motion dataset.
        The open-loop testing reveals that while the planning outcomes using our method do not always resemble the expert trajectory, they consistently outperform baseline approaches with improved safety, traveling efficiency, and riding comfort. The closed-loop testing further demonstrates the effectiveness of optimizing decisions and improving driving performance.    
        Ablation studies demonstrate that the initialization provided by the learning-based prediction module is essential for the convergence of the optimizer as well as the overall driving performance.
\end{abstract} 



\section{Introduction}    
\begin{figure}[!t]
        \centering
        \begin{tikzpicture}
            \node[anchor=center] {\includegraphics[width=1\linewidth]{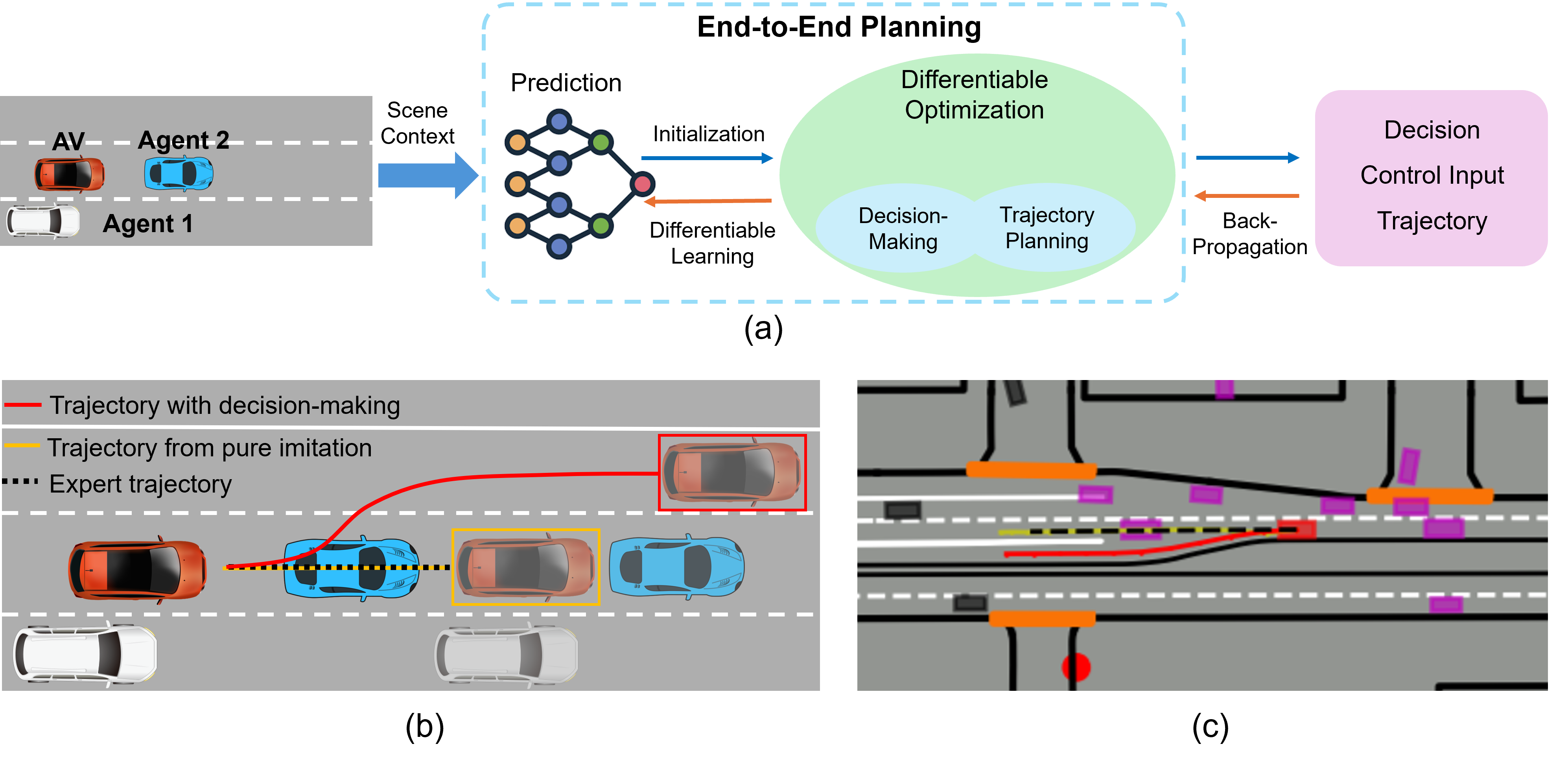}};
            \node[anchor=north west, text=black] at (0.2,1.5) {\scalebox{0.5}{$\mathop{minimize}\limits_{b,x,u}\ J(b,x,u)$} };
            \node[anchor=north west, text=black] at (2.2,1.6) {\scalebox{0.5}{$b^*,x^*,u^*$}};
    \end{tikzpicture}
    \caption{Illustration of our proposed approach. 
        (a) The end-to-end planning system leverages a modular architecture of prediction and differentiable optimization while maintaining end-to-end trainability.
        (b) In a multi-lane driving scenario, imitation learning methods typically produce a trajectory (yellow line) that closely aligns with the expert's path (black dotted line). In contrast, our method enables AV to choose the optimized lane for travel and generate a corresponding trajectory (red line) through decision-making, which enhances safety and efficiency by allowing the AV to navigate toward lanes free from obstacles in this scenario.
        (c) The effectiveness of the proposed method is demonstrated in our experiments. Our approach (red line) prioritizes safety and efficiency by dynamically selecting obstacle-free lanes, as compared to the planning outcome of~\cite{huang2023differentiable} that yields a trajectory that closely follows the expert's (yellow line).}
    \label{fig:fig1}
\end{figure}

    {A}{utonomous} driving (AD) has emerged as a critical innovation in modern transportation, revolutionizing the way that vehicles operate and interact~\cite{chib2023recent}.
    The conventional AD system typically follows a standalone design for individual tasks~\cite{martinez2019autonomous}. Recent research has highlighted that integrating these tasks within a unified framework fosters synergy among them, thereby enhancing the overall system performance~\cite{wenru2024tso}.
    In this context, the end-to-end formulation~\cite{{chen2024end}} for AD system represents a fully differentiable paradigm that integrates various tasks (i.e., perception, planning, and control) to optimize performance toward the overarching objectives of AD~\cite{{chen2022milestones}}. 
    Specifically, end-to-end planning\footnote{The term ``end-to-end planning systems" is distinct from ``end-to-end autonomous driving systems." End-to-end autonomous driving refers to fully differentiable programs that take raw sensor data as input and produce control actions as output, while end-to-end planning  refers to an integrated approach to prediction, decision-making, and planning that is typically characterized by a modular design.}~\cite{{hagedorn2023rethinking}}
    as shown in Fig.~\ref{fig:fig1} encompasses the integration of motion prediction, decision-making, and trajectory planning tasks within the AD system, leveraging the outputs or intermediate representations generated by the perception module. This framework effectively addresses the challenges faced by autonomous vehicle (AV) in interacting with complex traffic participants, enabling informed decision-making and the execution of safe actions in real time~\cite{{lecun2022path}}. Such capabilities are crucial for the success of AD tasks and achieving high-level autonomy~\cite{{guo2019safe}}. 
    This research theme has received increasing attention and bolstered by recent advancements in learning-based approaches~\cite{{zhu2021survey, ma2022local,zhou2024game}} and optimization-based approaches~\cite{{eiras2021two, huang2024integrated,liu2024improved}}.

    The learning-based approach, leveraging the powerful capabilities of deep learning networks, enables the AD system to adapt to a wide range of driving scenarios by effectively learning from extensive datasets collected from driving demonstrations~\cite{grigorescu2020survey}. 
    However, a significant limitation of this approach is the lack of interpretability, which arises from the inherently opaque structure of deep learning networks. 
    Furthermore, it is pertinent to note that the imitation learning (IL) paradigm~\cite{pan2020imitation} focuses on mimicking expert demonstrations. Consequently, the planning outcomes derived from this approach generally lack a comprehensive evaluation of optimal performance across critical dimensions, including safety, traveling efficiency, and riding comfort.
    In this regard, optimization-based methods provide a systematic framework for incorporating the notion of optimality mentioned above into AD tasks by clearly defining optimization objectives~\cite{lim2019hybrid},~\cite{ma2022alternating}. 
    However, the optimization process is often hindered by challenges such as the presence of local optima~\cite{schwarting2017safe} and difficulties in convergence~\cite{bonnans2006numerical}, diminishing its overall capacity to navigate complex and dynamic driving environments.
    In this sense, the hybrid approach has a recognized potential that allows the AD system to learn from large-scale datasets while utilizing an optimization process to guide the upper-stream learning process.
    However, in typical existing frameworks like~\cite{huang2023differentiable}, the decision-making process is typically overlooked, and planning is primarily confined to adhering to the expert trajectory. Essentially, this hinders the planning outcomes to closely resemble that of the expert demonstrations.

    With the above descriptions as a backdrop, it motivates us to address the capability to determine the desirable lane for travel and generate the corresponding trajectory within existing end-to-end planning frameworks. Specifically, we aim to achieve planning outcomes that extend beyond expert trajectories while demonstrating improved driving performance in terms of safety, traveling efficiency, and riding comfort.
    This becomes increasingly imperative with the emergence of mapless AD system~\cite{ort2019maplite} where the AV must autonomously select a lane from all available lanes to travel, compared with the HD-map-based approach where the decision is implicitly contained within the sequence of waypoints provided.
   Nevertheless, utilizing the optimization process (especially the decision-making ability) to guide end-to-end learning requires the optimizer to be differentiable, so that it can be compatible with other learning-based modules. However, formulating a decision-making task as a differentiable optimization problem presents two major challenges.
    First, the decision variables exhibit inherent discrete characteristics. Second, the vehicle can only select one option from a set of available choices, imposing an equality constraint. 
   This intrinsic nature of the decision-making problem necessitates innovative approaches to differentiable optimization that accommodate these discrete characteristics and equality constraints, ensuring compliance with these constraints throughout the learning process.

    In this study, we generalize the idea from \cite{huang2023differentiable} and propose an end-to-end planning framework that integrates motion prediction, decision-making, and trajectory planning tasks, with the aim of guiding the learning from expert demonstration with essential imperatives of safety, traveling efficiency, and riding comfort. The overview of this development is shown in Fig.~\ref{fig:fig1}. This key feature of decision-making capability, which distinguishes it from previous work, guides the end-to-end planning system toward improved driving performance. This capability is achieved through an integrated optimization problem that generates optimized planning outcomes, including both decisions and their corresponding trajectories.
    Essentially, the integrated optimizer is differentiable and is thus end-to-end trainable with the learning-based upstream modules (i.e., the prediction module in this study).
    Moreover, necessary constraints in the decision-making problem are considered in the problem formulation and enforced in the joint learning process.
    On top of that, a transformer-based neural network is utilized to predict trajectories for surrounding agents and initial plans including decision and trajectory for the AV that acts as the ego vehicle. These learned components act as initializations for the optimizer and play a crucial role in the convergence of the optimization process.  
    The ultimate outputs of the proposed framework include the joint prediction of future trajectories of agents, the optimized lane-selection decision, and the respective trajectory of AV.
    The proposed framework is trained on the Waymo Open Motion Dataset, which is based on the human driving experience from real-world driving scenarios. The driving performance of the proposed framework is examined in both open-loop and closed-looped testings, and the experiments show that our framework excels in diverse driving tasks.

    Our contributions are summarized as follows:
    \begin{itemize}
        \item We formulate the decision-making and trajectory planning tasks as a differentiable constrained nonlinear optimization problem, leveraging its differentiability for compatibility with a learning-based AD system. Specifically, the discrete nature and inherent constraint in the decision-making problem are addressed, and learned initialization for the decision variables is utilized to facilitate convergence in the optimization process.
        \item We propose an end-to-end planning framework where the learning-based prediction module is guided by the downstream optimizing objectives. Furthermore, the integration of decision-making capability enables the AV to make optimized decisions that transcend the imitation of expert demonstrations, which ultimately leads to enhanced performance in planning outcomes.
        \item The proposed framework is trained end-to-end and validated using the Waymo Open Motion Dataset. Our method outperforms the baseline methods in open-loop testing, while the optimized decision-making capability is showcased in the closed-loop simulation. The ablation study further highlights the critical role of the learned initialization in facilitating the convergence of the optimization process within the framework.
    \end{itemize}

    The organization of this paper is as follows: Section~\ref{sec:related} reviews research topics closely related to our study. Section~\ref{sec:framework} introduces the problem formulation regarding the proposed AD framework. Section~\ref{sec:method} presents the development of the differentiable optimizer and the joint training process in detail. Section~\ref{sec:exp} presents the results and discussions in the experiments conducted. Section~\ref{sec:con} summarizes our study and proposes future research directions. 
    
\section{Related Works} \label{sec:related}
    It is widely recognized that the different driving tasks in AD systems should collaboratively contribute to the overall planning outcomes, as highlighted by the planning-oriented approach~\cite{{hu2023planning}}.
    With the advancements of deep learning methods, there has been a surge in the integration of AD tasks through end-to-end driving system~\cite{jiang2023vad, anzalone2022end, le2022survey}. The end-to-end approach streamlines the architecture by consolidating all stages into a single model. However, it requires extensive and diverse driving data to effectively generalize across various scenarios~\cite{wang2024drive}.
    Moreover, the end-to-end paradigm often suffers from limited interpretability regarding the planned outcomes~\cite{chen2024ir}. 
    One potential effort to address interpretability involves providing the end-to-end system with enriched informative content.
    For instance, depth modality is employed in~\cite{{xiao2020multimodal}} to address the limitations of image-only input, and navigational commands, such as subgoal angles~\cite{{wang2019end}}, are utilized to improve prediction accuracy and reduce ambiguities at intersections. Furthermore,
    intermediate representations, including detections, predicted trajectories, and cost volumes, are also utilized in the end-to-end system to form an interpretable neural motion planner~\cite{zeng2019end}.
    The IL strategy also leverages the knowledge from experts such as model predictive control (MPC)~\cite{tagliabue2024efficient} and human feedback~\cite{yuan2024evolutionary} for the development of reliable deep neural network policy that can be effectively transferred to new domains.
    Nonetheless, as noted by~\cite{bansal2018chauffeurnet}, purely data-driven learning approaches are often inadequate for handling complex driving scenarios. The limitations inherent in this data-centric methodology can impede further improvements in the end-to-end system and negatively impact overall driving performance.

    In addition to unifying all driving stages into a neural network, optimization is also under extensive examination to coordinate these components for enhanced performance, along with innovations in structural design~\cite{{lang2024bev}}. Notably, differentiable optimization~\cite{{amos2018differentiable}} has been widely adopted in AD and robot learning, incorporating optimization problems into a learning-based framework by enabling backpropagation through gradient calculation. A prominent application of this approach is the differentiable kinematic model, which utilizes an optimization-based system to bridge the gap between planned trajectories and expert trajectories~\cite{{zhou2021exploring, cui2020deep}}. 
    The differentiable optimization process also provides a way to maintain the satisfaction of constraints along with task completion. LeTO~\cite{{{xu2024leto}}} integrates differentiable trajectory optimization into the IL framework that not only performs manipulation tasks but also complies with constraints in robotics. The algorithm proposed in~\cite{{donti2021dc3}} ensures the fulfillment of equality and inequality constraints through a completion and gradient-based correction procedure, with further extensions in~\cite{{diehl2022differentiable}} to IL settings for robots and AVs. 
    Another perspective on the application of differentiability involves creating a differentiable stack across various driving modules in AV, enabling end-to-end trainability while retaining modularity. This approach facilitates joint learning between motion prediction and motion planning modules in~\cite{{huang2023dtpp}}. DiffStack~\cite{{karkus2023diffstack}} presents a model-based method that integrates prediction, planning, and control into a joint optimization framework, utilizing gradients generated through the differentiable process.
   However, it is important to note that the decision-making task is generally overlooked in the studies mentioned above.
    In the absence of a decision-making process, AV planning typically generates trajectories based on pre-configured references, leading to schemes that merely imitate the trajectories of human drivers. 
    Our research specifically aims to address this gap in the existing literature by enabling AVs to make optimized decisions that extend beyond mere demonstrations, thereby generating corresponding trajectories that ultimately enhance the driving experience.

    \begin{figure*}[t]
        \centering
        \includegraphics[trim=0 0 0 0, clip, width=1\linewidth]{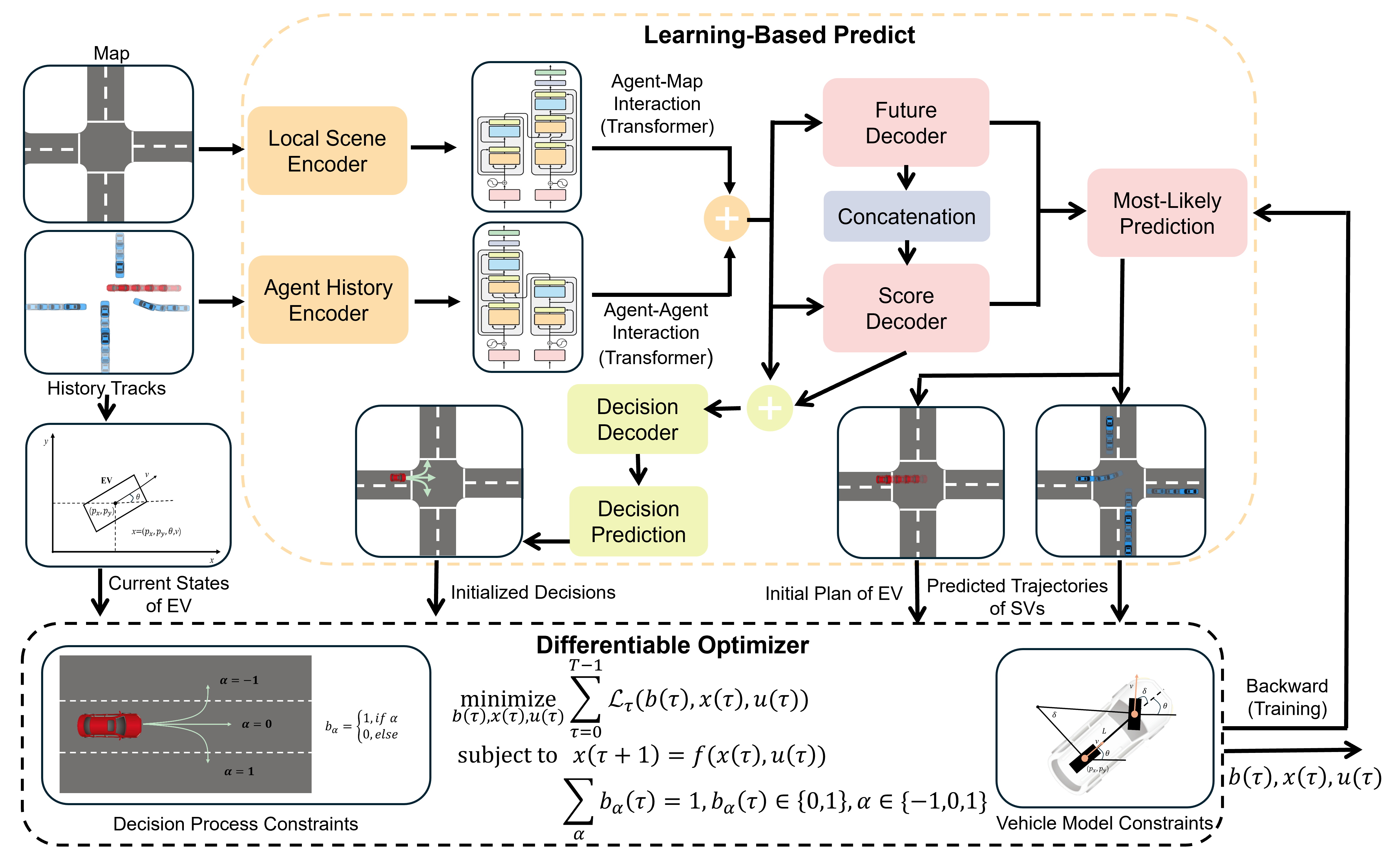}
        \caption{Pipeline of learning-based predictor and the differentiable optimizer for the integrated decision-making and trajectory planning tasks. The proposed AD framework is end-to-end trainable.}
        \label{fig:framework}
    \end{figure*}

\section{Problem Formulation}\label{sec:framework}
        As shown in Fig.~\ref{fig:framework}, the proposed framework consists of a learning-based predictor for future trajectories of the AV and other agents, and a differentiable optimizer for decision-making and trajectory planning problems.
        In the multi-lane driving scenario, assuming an accurate perception system, the inputs to the proposed AD framework consist of the historical trajectories of the AV and other agents, the current states of the AV, and vectorized maps that include available lanes and neighboring crosswalks.
        Within the proposed framework, the transformed-based predictor and the differentiable optimizer for decision-making and trajectory planning problems are trained jointly, which facilitates the generation of initial predictions and the optimization of the final planning outcome to be guided by overall driving performance.

     \subsection{Transformer-Based Motion Predictor} \label{sec:pred}
        The neural network predictor developed for this study employs a multi-modal approach to forecast future states within a dynamic environment featuring multiple agents.
        Taking the AV and agents as a cohort, the input data sources include the historical trajectories for this cohort, and map information of lanes and crosswalks, collectively denoted as the scene content.
        The encoders consist of the Local Scene Encoder to process spatial context derived from the map, and the Agent History Encoder to analyze the temporal context based on vehicles' current and past states.
        The features processed by both encoders are subsequently input into two distinct transformer-based interaction modules to model the AV's relationship with the environment and with neighboring agents.
        The outputs from these interaction modules are combined and fed into the decoders, consisting of \( K \) futures for this cohort, denoted as \( \{ \mathbf{F}_k | k = 1, 2, \ldots, K \} \), where each future is defined as \( \mathbf{F}_k = \{ f_k, s_k, d_k \} \).
        Here, the Future Decoder is responsible for generating the predicted trajectory $f_k$, which is a sequence of predicted state \( \mathbf{\hat{x}}^{0:N}_{1:T} \), with the superscript \( i \in \{0:N\} \) indicating the index of the vehicle, and the subscript referring to future time \( \tau \) over the next \( T \) timesteps. 
        Accompanying the Future Decoder, the Score Decoder evaluates the likelihood of each predicted trajectory to assess potential futures according to their probability, outputting the score value \( s_k \).
        Together, the Future Decoder and Score Decoder produce the most plausible outcomes.
        The outputs from the Future Decoder and Score Decoder, alongside previously extracted features, are concatenated to form a comprehensive feature set. This integrated data is then processed by the Decision Decoder, which generates a set of initialized decisions \( d_k \) for the AV.
        
        The outputs from the neural network predictor are passed to the optimizer, acting as initialization in the optimization formulation. Finally, the outputs of the proposed AD framework contain predicted trajectories for agents, optimized decisions and respective trajectories for the AV.
        
    \subsection{Problem Statement} \label{sec:decision}
        The decision-making problem defined in our study focuses on lane selection, whereby the AV determines which lane to travel in from a set of candidate lanes. 
        For clarity, we identify the leftmost lane as lane \( 1 \), increasing the lane ID by one from left to right. Accordingly, lane change maneuvers, including left lane change, lane keeping, and right lane change, are denoted by the discrete variable \( \alpha \in \{-1, 0, 1\} \), representing each maneuver respectively. This discrete variable \( \alpha \) is further transformed into a binary variable \( b_{\alpha} \), where \( b_{\alpha} = 1 \) if the respective maneuver \( \alpha \) is selected, and \( b_{\alpha} = 0 \) otherwise. 
        This decision subsequently informs the reference lane and velocity profiles to be tracked for the trajectory planning problem.
        Consistent with the systematic design of the proposed AD framework, the decision-making outcome is initially generated by the learning-based prediction model using a SoftMax function. The inherent constraints in the decision-making problem are then incorporated during the optimization stage.
        Considering that the AV can select only one lane at any given time, we impose the constraint that the sum of the values of \( b_{\alpha} \) across all possible values of \( \alpha \) must equal one:
        \begin{align}
        \sum_{\alpha} b_{\alpha}(\tau) = 1, \quad b_{\alpha}(\tau) \in \{0, 1\}, \quad \alpha \in \{-1, 0, 1\}.
        \end{align}
        
        With this clarification, we introduce the concept of a target lane associated with a decision. The starting lane for the AV is denoted as lane \( \sigma \); thus, the target lane corresponding to decision \( \alpha \) is lane \( \sigma + \alpha \).
        
    \subsection{Vehicle Model} \label{sec:vm}
        A reliable vehicle model is needed in the generation of feasible trajectories for the AV. Essentially, we need to convert the control inputs into trajectories in calculating certain costs in the optimization objective function using the vehicle model. To maintain differentiability for the computation of gradients and Jacobians within the optimization problem, we employ a kinematic bicycle model~\cite{huang2023differentiable}.         
        The state vector for the AV is defined as $x=(p_x, p_y, \theta, v)$, where $p_x$ and $p_y$ are the positions of the center point of the vehicle on the $X$ (longitudinal) and $Y$ (lateral) directions, respectively; $\theta$ is the heading angle of the vehicle, and $v$ is velocity. The control input vector $u=(a, \delta)$ consists of the acceleration and steering angle. 
        With \( \Delta \tau \) as the time step and \( L \) being the wheelbase of the vehicle, we have the following expression for the nonlinear discrete vehicle model:
        \begin{equation}
        \begin{aligned} 
        \label{eq:dynamics}
        p_x(\tau+1) &= p_x(\tau) + v(\tau) \cos(\theta(\tau)) \Delta \tau, \\
        p_y(\tau+1) &= p_y(\tau) + v(\tau) \sin(\theta(\tau)) \Delta \tau, \\
        \theta(\tau+1) &= \theta(\tau) + \frac{v(\tau)}{L} \tan(\delta(\tau)) \Delta \tau, \\
        v(\tau+1) &= v(\tau) + a(\tau) \Delta \tau.
        \end{aligned} 
        \end{equation}

    \subsection{Optimization of Decision-Making and Trajectory Planning} \label{sec:nlp}
    Based on the definitions provided, the initial decisions and plans for the AV, along with the predicted trajectories of surrounding agents, are integrated into a nonlinear programming problem:
        \begin{subequations}
        \begin{align} 
             \displaystyle\operatorname*{minimize}_{b(\tau), x(\tau),u(\tau)}\quad
            &  \sum_{\tau=0}^{T-1} \ell_\tau\big(b(\tau), x(\tau),u(\tau)\big) \label{eq:4a} \\
            \operatorname*{subject\ to}\quad
            &x(\tau+1)=f\big(x(\tau),u(\tau)\big)  \label{eq:4b},  \\
            &\sum_{\alpha}b_{\alpha}(\tau) = 1, \label{eq:alpha1}\\
            & b_{\alpha}(\tau) \in \big\{0,1\}, \label{eq:alpha2}\\
            & \alpha \in \{{-1},0,1\}, \label{eq:alpha3}  \\
            &\tau=0,1,\dotsm, T-1. \nonumber
        \end{align}
        \end{subequations}

This problem aims to minimize the costs associated with the AV's state $x(\tau)$, control inputs $u(\tau)$, and decisions $b(\tau)$, ultimately leading to the determination of the optimal values for these variables.
        This problem enables the optimization of decision-making and trajectory planning jointly, where~\eqref{eq:4a} integrates the decision-making and trajectory planning objectives, covering a variety of factors based on safety, traveling efficiency, and riding comfort. The constraints~\eqref{eq:4b} refer to the vehicle model, and~\eqref{eq:alpha1} to~\eqref{eq:alpha3} are the constraints inherent to the decision-making problem.
        We further transform this nonlinear optimization problem into differentiable optimization in the next section. 
        In this way, the gradient from the optimization problem can be back-propagated, and the whole AD framework is end-to-end trainable to allow the learning components within the framework to be supervised by the ultimate driving result.

\section{Methodology} \label{sec:method}
    \subsection{Integrated Optimization Objective} ~\label{sec:int_J}    
        At a high level, we regard the differentiable optimizer as the primary component of our framework.
        Building upon our previous work~\cite{wenru2024tso}, we present in this subsection the formulation of the objective function within the integrated decision-making and trajectory planning optimization, along with the necessary transformations to ensure the differentiability of the optimization problem.      

        \subsubsection{Position Tracking}
            The position tracking cost is defined as the distance between the AV's actual positions and the lane center of each of the available lanes. The decision result $\alpha$ directly determines the lane for the AV to drive on, and it is important for the vehicle to adhere to the geometric structure of the lane. Thus, we devise the position tracking cost directly in association with each decision result $\alpha$:
            \begin{align}\label{}
                \ell_{tr,\alpha}(\tau) &= w_{tr,x} (p_{x}(\tau) - p_{x,\sigma+\alpha}^r(\tau))^2  \\
                &+ w_{tr,y} (p_{y}(\tau) - p_{y,\sigma+\alpha}^r(\tau))^2, \nonumber
            \end{align}
            where $p_{x,\sigma+\alpha}^r$ and $p_{y,\sigma+\alpha}^r$ are the $X$ and $Y$ coordinates of the center of the target lane $\sigma+\alpha$ with respect to the decision $\alpha$, and the positive constants $w_{tr,x}$ and $w_{tr,y}$ are the weighting coefficients.

            Taking into consideration all possible values of the decision $\alpha$, we have the position tracking cost for the integrated problem of decision-making and trajectory planning:
            \begin{align} \label{eq:tr}
               \ell_{tr}(\tau) 
               &= b_{-1}(\tau) \ell_{tr,-1}(\tau) 
               + b_{0}(\tau) \ell_{tr,0}(\tau) \\
               &+ b_{1}(\tau) \ell_{tr,1}(\tau). \nonumber
            \end{align}

        \subsubsection{Decision Safety}
            We focus on the discussion on vehicles in this section, while other traffic participants such as pedestrians and cyclists can be attempted in a similar fashion.
            The decisions regarding lane-keeping and lane-changing inherently generate safety costs in both longitudinal and lateral dimensions. 
            We begin our analysis with the longitudinal safety cost, which pertains to leading vehicles (LVs), specifically the vehicles ahead of the AV. This cost is crucial for informing both lane-keeping and lane-changing decisions. When an LV occupies the target lane, the corresponding lane-selection decision must account for a safety cost, as the AV must maintain a safe longitudinal distance from the LV to mitigate collision risks.

            Considering the decision $\alpha$ and the respective target lane $\sigma+\alpha$, we denote the positions of the LV on the target lane as $p_{x,\sigma+\alpha}^{LV}$ and $p_{y,\sigma+\alpha}^{LV}$, and the longitudinal velocity of the LV as $v_{x,\sigma+\alpha}^{LV}$.
            The cost term $\ell_{d-lon,\alpha}$ is defined to describe the cost arising from the interaction between the AV and the LV for the respective decision $\alpha$, and we have
            \begin{align}\label{eq:lon-cost}
                \ell_{d-lon,\alpha}(\tau) &=  w_{v-lon} \xi_{\sigma+\alpha}^{LV}(\tau) ( v_{x,\sigma+\alpha}^{LV}(\tau)-v_{x}(\tau))^2 && \\ \nonumber & + \frac{w_{d-lon}}{(\Delta d_{\sigma+\alpha}^{LV}(\tau))^2+\varepsilon^2}, &&
            \end{align}
            where the positive constants $w_{v-lon}$ and $w_{d-lon}$ are weighting coefficients, and a small nonzero constant $\varepsilon$ is added to the denominator for the consideration of numerical stability by preventing division by zero.
            
            In the above definition, $\Delta d_{\sigma+\alpha}^{LV}(\tau)$ is a safety distance from the LV on the respective target lane $\sigma+\alpha$ regarding a decision $\alpha$:
            \begin{align}\label{eq:lon-d}
                &\Delta d_{\sigma+\alpha}^{LV}(\tau)\\ \nonumber
                &= \sqrt{(p_{x,\sigma+\alpha}^{LV}(\tau)-p_{x}(\tau))^2+(p_{y,\sigma+\alpha}^{LV}(\tau)-p_{y}(\tau))^2} - l,
            \end{align}
            with $l$ being the length of the vehicle.
            
            We further specify $\xi_{\sigma+\alpha}^{LV}(\tau)$ as an indicator function that only calculates the velocity difference if the LV is driving slower than the AV, which makes the safety cost in the longitudinal direction more reasonable because the safe distance is only relevant when there is a slow-moving vehicle ahead:
            \begin{align}\label{eq:lon-sgn}
                \xi_{\sigma+\alpha}^{LV}(\tau)= \left\{ \begin{aligned}
                    &1\quad\quad \text{if} \ v_{x,\sigma+\alpha}^{LV}(\tau)-v_{x}(\tau)<0 \\
                    &0 \quad\quad \text{if} \ v_{x,\sigma+\alpha}^{LV}(\tau)-v_{x}(\tau) \geq0, \\
                \end{aligned}
                \right.
            \end{align}

            By considering all possible values of the decision $\alpha$, we have the safety cost associated with LVs as the following:
            \begin{flalign} \label{eq:s_lv}
               \ell_{s\_LV}(\tau) &= b_{-1}(\tau) \ell_{d-lon, -1}(\tau) + b_{0}(\tau) \ell_{d-lon, 0}(\tau)  \\
               &+ b_{1}(\tau) \ell_{d-lon, 1}(\tau).  \nonumber
            \end{flalign}

            Similarly, we examine the safety cost in the lateral direction, which is integral to the lane-changing decision and relates to the neighbor vehicles (NVs), specifically, the vehicles approaching from behind in adjacent lanes. This safety cost is defined as follows:
            \begin{flalign}\label{eq:lat-cost}
                \ell_{d-lat,\alpha}(\tau) &= w_{v-lat} \xi_{\sigma+\alpha}^{NV}(\tau) ( v_{x,\sigma+\alpha}^{NV}(\tau)-v_{x}(\tau))^2 && \\ & + \frac{w_{d-lat}}{(\Delta d_{\sigma+\alpha}^{NV}(\tau))^2+\varepsilon^2}, \nonumber
            \end{flalign}
            \begin{align}\label{}
                &\Delta d_{\sigma+\alpha}^{NV}(\tau) \\ 
                & = \sqrt{(p_{x,\sigma+\alpha}^{NV}(\tau)-p_{x}(\tau))^2+(p_{y,\sigma+\alpha}^{NV}(\tau)-p_{y}(\tau))^2} - l, \nonumber
            \end{align}
            \begin{align}\label{eq:lat-sgn}
                \xi_{\sigma+\alpha}^{NV}(\tau)  = \left\{ \begin{aligned}
                    &1\quad\quad \text{if} \ v_{x,\sigma+\alpha}^{NV}(\tau) - v_{x}(\tau) >0\\
                    &0 \quad\quad \text{if} \ v_{x,\sigma+\alpha}^{NV}(\tau) - v_{x}(\tau) \leq0,\\
                \end{aligned}
                \right.
            \end{align}		

            In this context, $v_{x,\sigma+\alpha}^{NV}$ denotes the velocity of the NV on the target lane, while $p_{x,\sigma+\alpha}^{NV}$ and $p_{y,\sigma+\alpha}^{NV}$ represent the positional coordinates of the NV within the target lane. The positive constants $w_{v-lat}$ and $w_{d-lat}$ serve as weighting coefficients that determine the relative importance of velocity and distance in the safety assessment. Additionally, $\xi^{NV}$ is an indicator function that applies a penalty for velocity differences only when the NV is traveling faster than the AV. By incorporating penalties associated with rapidly approaching vehicles from adjacent lanes, the safety of the lane-changing process is further reinforced.

            Taking the lane changing decision to the left and right adjacent lanes into account, the safety cost considering the NVs are defined as the following:     
            \begin{flalign} \label{eq:s_nv}
               \ell_{s\_NV}(\tau) = b_{-1}(\tau) \ell_{d-lat, -1}(\tau) + b_{1}(\tau) \ell_{d-lat, 1}(\tau). \yesnumber 
            \end{flalign}

        \subsubsection{Traveling Efficiency}
            The AV seeks to achieve higher traveling efficiency while adhering to the designated speed limit for the lane in which it is traveling. Similar to the rationale for the position tracking cost, the AV depends on the optimization process to guide its lane-selection decision which gives rise to the associated speed limit. Consequently, we introduce the following cost term to link the reference speed, defined as the speed limit for the target lane, with each decision $\alpha$:
            
            \begin{align}\label{}
                \ell_{eff,\alpha}(\tau)= w_{velo} (v_{x}(\tau) - v_{x,\sigma+\alpha}^{limit}(\tau))^2,
            \end{align}
            where {$v_{x,\sigma+\alpha}^{limit}$} is the speed limit of the target lane, and $w_{velo}$ is the weighting parameter.
    
            Considering  each decision option, the traveling efficiency cost in the integrated objective function are defined as the following:
            \begin{flalign} \label{eq:eff}
               \ell_{eff}(\tau) & = b_{-1}(\tau) \ell_{eff,-1}(\tau) 
                 + b_{0}(\tau) \ell_{eff,0}(\tau) \\
                &+ b_{1}(\tau) \ell_{eff,1}(\tau). \nonumber
            \end{flalign}

        \subsubsection{Riding Comfort}
            The riding comfort cost is based on the acceleration $a$ and the steering angle $\delta$: 
            \begin{align}\label{}
                \ell_{rc}(\tau)
                = w_{rc,1} (a(\tau))^2 + w_{rc,2} (\delta(\tau))^2,
            \end{align}
            where the positive constants $w_{rc,1}$ and $w_{rc,2}$ are weighting coefficients. 
        
        \subsubsection{Driving Compliance}
            The cost terms considered so far are soft costs, which are flexible penalty terms incorporated into the cost function to encourage desired behaviors without strictly enforcing compliance. 
            There exist non-negotiable constraints that must be satisfied to ensure lawful operation in safe driving, including hard constraints for collision avoidance and compliance with traffic signals. As we formulate our problem as a differentiable nonlinear least-squares optimization, we convert these hard constraints into soft penalty terms within the cost function. These penalty terms are assigned large cost weights that remain fixed and are not subject to adjustment during the cost weight learning process.

            We borrow the cost definitions from~\cite{huang2023differentiable} for collision avoidance and traffic light compliance.  
            The collision avoidance cost is calculated in the Frenet frame, which allows for effective tracking of the safe distance between the AV and the relevant agent. The safety cost reflects a substantial penalty whenever the distance between the AV and the agent falls below the established safe limit. Therefore, the hinge loss is employed to impose a significant penalty for violating the safe distance threshold:
            \begin{equation}
            \label{eq8}
            \ell_{safety}(\tau) = \begin{cases} w_{safe} (\epsilon - d_{safe}), & d_{safe} \leq \epsilon \\ 0, & \text{otherwise} \end{cases}.
            \end{equation}
            
            Note that the weighting parameter $w_{safe}$ is a constant pre-configured for enforcing the hard constraint on collision avoidance. The constant $\epsilon$ is the minimum safety distance requirement, which is defined as the sum of the lengths of two agents and a safety gap. 
            The safe distance is calculated in Frenet frame to save computation as only the interactive agent is considered in the calculation:
            \begin{equation}
            \label{eq7}
            d_{safe} = \min_{i} \parallel p_{\tau} - p_{\tau}^i \parallel_2,
            \end{equation}
            where $p_{\tau}^i$ is the predicted position of the interactive agent $i$ at future timestep $\tau$.

            The cost associated with violating traffic signals is similarly computed using hinge loss:
            \begin{equation}
            \label{eq6}
            \ell_{traffic}(\tau) = \begin{cases} w_{stop} (d_{stop} - p_{stop}), & d_{stop} \geq p_{stop} \\ 0, & \text{otherwise} \end{cases}.
            \end{equation}
                        
            Here, the weighting parameter $w_{stop}$ is the weight for the hard constraint of traffic light compliance. We calculate the running distance of AV at the current speed for one timestep and represent it as $d_{stop}$. The stop line position is denoted as $p_{stop}$.
            This implies that if the AV crosses the stop line at a red light, a substantial penalty will be incurred, thereby incentivizing the AV to halt appropriately near the stopping point.

        \subsubsection{Integer Variable}
            Another non-negotiable constraints are the integer constraint and equality constraint inherent in the decision-making problem, which are defined as~\eqref{eq:binary} and ~\eqref{eq:sumb_1}, respectively:
            \begin{align}\label{eq:binary}
            \ell_{binary}(\tau) = w_{bi} \max(0, b_{\alpha}(b_{\alpha} - 1)),
            \end{align}    
            \begin{align}\label{eq:sumb_1}
            \ell_{equality}(\tau) = w_{eq} \max(0, b_{-1} + b_{0} + b_{1} - 1)).
            \end{align}

            The weights $w_{bi}$ and $w_{eq}$ are assigned with large values for the compliance of the hard constraints during the training process.

            To render the optimization problem differentiable, it is necessary to relax the decision-making variables from binary to continuous values within the range of zero to one. This relaxation is achieved through~\eqref{eq:binary} and~\eqref{eq:sumb_1}, which allow for the continuous representation of these variables while ensuring that the associated constraints~\eqref{eq:alpha2} and~\eqref{eq:alpha1} in the decision-making process are still respected.

        In summary, the integrated objective function is defined as follows:
        \begin{flalign} \label{eq:obj_basis}
           & \ell_\tau\big(b(\tau),x(\tau),u(\tau)\big) \\
           & =  \ell_{tr}(\tau) + \ell_{s\_LV}(\tau) + \ell_{s\_NV}(\tau)  + \ell_{eff}(\tau) + \ell_{rc}(\tau) \nonumber \\ 
           & + \ell_{safety}(\tau) + \ell_{traffic}(\tau) + \ell_{binary}(\tau) + \ell_{equality}(\tau). \nonumber
        \end{flalign}

        \begin{remark}
        The objective function in this section comprehensively addresses the decision-making and trajectory planning process, incorporating the notion of optimality into the planning outcome, rather than directly imitating human ground truth. 
        Decision is typically implicitly embedded in human demonstrations, reflecting a mixture of human preferences, intentions, and external mandatory commands (such as navigation), which are not directly available in the dataset.
        By considering all potential decision values in ~\eqref{eq:tr}, ~\eqref{eq:s_lv}, ~\eqref{eq:s_nv}, and ~\eqref{eq:eff}, we define the optimization objective from the downstream that guides the learning process, focusing on tracking, safety, and travel efficiency. 
        Additionally, factors such as riding comfort and driving compliance, which are important throughout the entire driving experience, are modeled independently of the chosen decision maneuver. The inherent constraints related to the decision-making variables are also integrated into the objective function.
        \end{remark}

    \subsection{Joint Training Process}
        From a holistic perspective, the structure of our proposed framework is a bilevel optimization setup~\cite{pineda2022theseus} as described below:
        \begin{equation}
        \begin{aligned}
          &\text{Inner loop:\;} \theta^\star(\phi) = \operatorname*{argmin}_\theta J(\theta; \phi), \\
          &\text{Outer loop:\;} \phi^\star = \operatorname*{argmin}_{\phi} L ( \theta^*(\phi) ).
          \label{eq:dnls}
        \end{aligned}
        \end{equation}

        The inner loop represents the optimization problem formulated in Section~\ref{sec:nlp} using~\eqref{eq:4a} to~\eqref{eq:alpha3}, where the optimization variable $\theta$ pertains to the lane-selection decision and the corresponding trajectories in our study. 
        By utilizing the objective function for integrated decision-making and trajectory planning as outlined in Section~\ref{sec:int_J}, this nonlinear programming problem can be transformed into a nonlinear least squares problem, which can be effectively addressed by existing differentiable optimizers. 
        The outer loop is the learning-based predictor, a neural network parameterized by $\phi$, that provides initialization on the decision and the corresponding trajectories in our study.
        The inner-loop objective $J$ is part of the outer loss $L$ that facilitates the end-to-end training.

        Our objective is to establish an end-to-end trainable framework that concurrently learns the parameters $\phi$ and solves the optimization variable $\theta$. 
        Regarding the inner loop, we solve the nonlinear least squares problem through the Gauss-Newton algorithm~\cite{huang2023decentralized}, which applies linearization on the nonlinear system iteratively and uses the Jacobian of residuals to converge to a solution. 
        A step size $\beta$ in the range of zero to one is applied for each update to maintain stability in the optimization process.
        The inner loop optimization provides the solution $\theta^\star$, along with the gradient $\partial\theta^\star/\partial\phi$ using this differentiable optimization procedure.
        The neural network in the upstream predictor then applies standard gradient descent on this gradient to update its parameter $\phi$.
        By integrating the differentiable optimization process with the standard deep-learning network, we have developed an end-to-end learning framework that enhances the consistency between the learned parameters $\phi$ in the upstream module and the planning outcomes produced by the downstream optimization process.

        Given that all operations within the framework are differentiable, we can effectively train the entire system using real-world driving data. We design a comprehensive loss function that encompasses the various tasks integral to the proposed framework. This includes prediction loss and score loss for the prediction task, decision loss for the decision-making task, as well as planning loss and imitation loss for the trajectory planning task.
    
        \begin{equation} \label{eq:loss_train}
        \begin{split}
        \mathcal{L} = & \lambda_1 \mathcal{L}_{prediction} + \lambda_2 \mathcal{L}_{score} + \lambda_3 \mathcal{L}_{decision} \\
        & + \lambda_4 \mathcal{L}_{planning} + \lambda_5 \mathcal{L}_{imitation},
        \end{split}
        \end{equation}
        where $\lambda_1$, $\lambda_2$, $\lambda_3$, $\lambda_4$, and $\lambda_5$ are the loss weights in the range of zero to one to scale the components in the loss function.

        The prediction loss is formulated for the neural network predictor, with the primary objective of guiding the upstream prediction module in generating predicted trajectories that act as initialization for the subsequent optimization process. Ground truth trajectories, as demonstrated by human drivers, provide a reliable reference for the prediction module. Specifically, the prediction loss for the AV is designed to measure the alignment of the predicted trajectory $f^k_i$ for each vehicle indexed $i$ with the respective ground truth \(\eta_i^{gt}\).
        Considering the multiple vehicles in the prediction task as a cohort, we identify the future with the smallest sum of displacement errors for all vehicles across $K$ possible future trajectories. The best-predicted future is denoted as $\hat k$, with the following definition:
        \begin{equation}
        \hat k = \arg \min_k \sum_i \parallel f^k_i - \eta_i^{gt}\parallel_2.
        \end{equation}

        The prediction loss is formally defined as follows:
        \begin{equation}
        \mathcal{L}_{\text{prediction}} = \sum_{k=1}^K \mathbbm{1}(k=\hat{k}) \sum_{i=1}^N \text{smooth} \, L_1 (f^k_i - \eta_i^{gt}),
        \end{equation}
        where \(\mathbbm{1}\) is the indicator function.

        Accompanying the prediction loss is the score loss, which quantifies the alignment between the predicted future with the highest score and the best-predicted future. It utilizes a cross-entropy loss with the following definition:
        \begin{equation}
        \mathcal{L}_{\text{score}} = - \sum_{k=1}^K \mathbbm{1}(k=\hat{k}) \log s_k,
        \end{equation}
        where \(s_k\) denotes the score value associated with the predicted trajectory \(k\). This formulation captures the notion that the score loss penalizes the divergence between the predicted probabilities and the expected outcomes, thereby reinforcing the model's ability to prioritize the most likely trajectories.


        In addition to generating possible future trajectories and associated scores, the prediction module also provides an initialized solution for lane-selection decision.
        The decision loss leverages the optimization process to guide the learning of decision-making, which is essentially a cross-entropy loss measuring the alignment of the learned decision from the predictor with the optimized decision.

        The decision loss is expressed as follows:
        \begin{equation}
        \mathcal{L}_{\text{decision}} = -\sum_{\alpha \in \{-1, 0, 1\}} \mathbbm{1} (\alpha = \alpha^*) \log({d}_{\alpha}), 
        \end{equation}
        where \({d}_{\alpha}\) is the predicted probability of selecting lane \(\alpha\), which acts as the initialized decision in the proposed AD framework. The indicator function matches the initialized decision ${\alpha}$ by the predictor with the optimizer decision $\alpha^*$ to effectively penalize discrepancies between the predicted and the optimized lane-selection decisions, thereby enhancing the model's decision-making capability.


        The trajectory planning task within the proposed AD framework is fundamentally an optimization problem. During the training process, two sources of error must be considered to evaluate the quality of the planned trajectories. The first is the planning loss, denoted by $\mathcal{L}_{\text{planning}}$, which represents the total squared error from the trajectory optimization. This loss aligns with minimizing the optimization objective, promoting smooth trajectories while avoiding obstacles. The second source of error is the imitation loss, which leverages the ground truth trajectory to guide the trajectory generated by the optimizer. The definition of the imitation loss is as follows:
        \begin{equation}
        \mathcal{L}_{\text{imitation}} = \text{smooth}{L_1} (\eta_{AV} - \eta_{AV}^{gt}).  
        \end{equation}
        
        Additionally, we utilize a pretraining phase to allow the prediction module to learn from data and provide reasonable initialization from scratch. It is important to note that the optimizer is not accessible during pretraining, and the planning results are not available. Consequently, learning in the pretraining phase relies solely on ground truth data from human drivers. The pretraining loss only considers the prediction loss and score loss from the neural network predictor, without involving the final outputs by the optimizer. The pretraining loss is defined as follows: 
        \begin{equation}
        \label{loss_pretrain}
        \mathcal{L}_{\text{pretrain}} = \lambda_1 \mathcal{L}_{\text{prediction}} + \lambda_2 \mathcal{L}_{\text{score}}.
        \end{equation}
        




\begin{algorithm}[t]
    \caption{Joint Training of Proposed Framework}  \label{Alg1}
    \begin{algorithmic} [1]
        \State \textbf{Input:} {Neural network for predictor $\mathcal{N}_\phi$, differentiable optimizer with optimizing variables $\theta$, integrated objective $J$ of initial cost function weights $\{\omega_i\}$}
        \State Prepare scene content with history tracks for vehicles and vectorized map
        \State Locate the AV to the nearest lane
        \State Attain the adjacent left and right lane if applicable
        \State Predict the joint trajectories for the cohort of vehicles with multiple possible futures, with each future consisting of the trajectory and associated probability
        \State For AV, predict the initialized decisions on selecting each lane from the available lanes in a three-lane configuration
        \State Calculate prediction loss $\mathcal{L}_{prediction}$ and score loss $\mathcal{L}_{score}$
        \State Initialize the differentiable optimizer with the most-likely predicted trajectories for agents, the initialized decision and trajectory for AV
        \State Solve the differentiable planner for the decision-making result and the respective trajectory
        \State Calculate decision loss $\mathcal{L}_{decision}$, imitation loss $\mathcal{L}_{imitation}$, and planning loss $\mathcal{L}_{planning}$
        \State Calculate total loss $\mathcal{L}$ according to ~\eqref{eq:loss_train}
        \State Backpropagate loss and calculate gradients with respect to $\theta$ and $\{\omega_i\}$
        \State Update $\phi$ and $\{\omega_i\}$ using Adam optimizer
        \State \textbf{Output:}  {Predicted trajectories  \( \mathbf{\hat{x}}^{0:N}_{1:T} \), optimized decisions $d^*_{\alpha}$, and the respective trajectories $\eta_{AV}$}
    \end{algorithmic}
\end{algorithm}

\section{Experiments} \label{sec:exp}
    The proposed AD framework integrates both data-driven and optimization-based methodologies. 
    The learnable components within the framework are trained using the Waymo Open Motion Dataset~\cite{sun2020scalability}, and the differentiable optimization is conducted with the support of Theseus~\cite{pineda2022theseus}.
    Note that it is a comprehensive large-scale dataset collected from human driving, encompassing a variety of typical urban driving scenarios and is particularly recognized for its complexity, diversity, and interactivity, which are essential for AD tasks. The dataset comprises 1,000 files, representing a total of 103,354 unique driving scenes, with each data file formatted as a 20-second video clip.
    In preparing the training dataset, we adhere to the data processing pipeline outlined by ~\cite{huang2023differentiable}. This process converts the original clips into frames that encapsulate the current states of AV and agents, the trajectories by human drivers, and detailed map elements, including lanes, crosswalks, traffic lights, and speed limits. A notable modification in our processed data is the absence of a reference route; instead, we determine the current lane of the AV by locating it to the nearest lane and infer the adjacent left and right lanes accordingly.
    From the dataset, we randomly sample 10\% (i.e., 100 data files) and filter out irrelevant scenarios, such as parking lots, one-lane roads, and instances of unnecessary halting, resulting in a total of 88,123 frames. We allocate 90\% of these frames for training, leaving 10\% for validation, which corresponds to 79,011 frames for training and 9,112 frames for validation.

    Referring to~\cite{huang2023differentiable},
    experiments are conducted in both open loop and closed loop, incorporating comparative study (Section~\ref{sec:comparative}) and ablation study (Section~\ref{sec:ablation}) to validate the efficacy and robustness of the proposed method. 
    The planned trajectory is compared with the ground truth in the open-loop testing, without actual execution of the planned results.
    The closed-loop testing employs a log-replay simulation utilizing the Waymo Open Motion Dataset, wherein the AV makes decisions and executes the first action from the planned trajectory, while the agent replays their recorded trajectories from the dataset.
    A summary of the hyperparameters utilized for the prediction network, as well as those for optimization during the training and inference stages, is provided in Table~\ref{tab:parameter}.
    
    \begin{table}[t]
      \centering
      \caption{hyperparameters of the main modules during training and inference process}
      \resizebox{1.0\linewidth}{!}{
        \begin{tabular}{ccc}
        \toprule
        \multicolumn{1}{c}{Module} & \multicolumn{1}{c}{Parameter} & \multicolumn{1}{c}{Value} \\
        \midrule
        \multicolumn{1}{c}{Prediction} & Number of neighbors $N$ & 10 \\
              & Number of predicted future $K$ & 3 \\
              & Historical timesteps $H$ & 20 \\
              & Future timesteps $T$ & 50 \\
        \midrule
        \multicolumn{1}{c}{Optimization (Training)} 
              & {Step size} $\beta$ & 0.4 \\
              & Planning iterations                      & 2 \\
              & Initial learning rate                    & 2.00e-04 \\
              & Total training epochs                    & 20 \\
              & Pretraining epochs                       & 3 \\
              & Batch size                               & 32 \\
              & Weight for prediction loss $\lambda_1$ & 0.5 \\
              & Weight for score loss $\lambda_2$ & 1 \\
              & Weight for decision loss  $\lambda_3$ & 1 \\
              & Weight for imitation loss $\lambda_4$ & 1 \\
              & Weight for planning cost  $\lambda_5$ & 0.1 \\
        \midrule
        \multicolumn{1}{c}{Optimization (Inference)} & {Step size} $\beta$ & 0.5 \\
              & Planning iterations                      & 10 \\
        \bottomrule
        \end{tabular}%
        }\label{tab:parameter}%
    \end{table}%

    \subsection{Evaluation Metrics}
    The open-loop testing emphasizes the similarity between the planned outcomes and the ground truth by human drivers, whereas closed-loop testing facilitates the actual execution of the planned actions, demonstrating the model's responsiveness to the driving environment. Despite the differences in testing modes, we develop evaluation metrics in aspects of safety, traveling efficiency, and comfort to provide a comprehensive assessment of the ultimate driving performance:
    \begin{enumerate}
        \item Safety: The collision rate serves as a bottom line of safety metric, with collisions recorded whenever the AV collides with agents at any time step during the simulation. Additionally, the safety index~\cite{wu2023integrated} is computed as the relative distance between the AV and the most relevant agent, normalized by the AV's traveling speed. A higher safety index indicates a greater safety margin between the AV and agents based on the AV's current speed. In open-loop testing, we also evaluate the off-route rate, which indicates deviations from the planned trajectory relative to the ground truth.
        \item Traveling Efficiency: Longitudinal progress in meters and average speed are calculated to reflect traveling efficiency.
        \item Riding Comfort: Average longitudinal and lateral accelerations are measured to provide a quantitative assessment of riding comfort along the planned trajectory.
        \item Planning Accuracy: The similarity between the planned trajectory and human trajectories is critical in open-loop testing. We calculate position errors between the planned trajectory and ground truth at intervals of 1, 3, and 5 seconds. This metric is exclusive to open-loop testing.
    \end{enumerate}

    \subsection{Comparative Study} \label{sec:comparative}
    We compare the performance of the proposed framework in both open-loop and closed-loop tests with three alternative models to comprehensively evaluate the efficacy of our proposed methodology.
    The same input data, network structure, hyperparameters, and output format as the proposed method are utilized in the implementation of alternative methods to ensure comparability of the testing results.
        \begin{enumerate}
            \item {Vanilla IL}: This approach trains a neural network to output the joint forecasting of agents, as well as the decision and trajectory for the AV. 
            The vanilla IL establishes a baseline for the capability of a learning-based method in achieving the desired tasks of prediction, decision-making, and trajectory planning             
            \item {IL + OPT}: We consider a stacked architecture of the learning-based prediction and the optimization for decision-making and trajectory planning. The neural network as the predictor is trained without the joint learning of the differentiable optimizer. The outputs from the predictor, containing the joint forecasting of agents and the initialization on decision and trajectory for the AV, are passed to the integrated decision-making and trajectory planning optimization for further refinement.
            Unlike the proposed method, which is end-to-end trainable, this configuration disrupts the linkage between learning-based prediction and optimization-based decision-making and trajectory planning tasks.
            The comparison aims to highlight the significance of the integrated learning structure in the proposed framework.
            \item {DIPP}: We choose to compare the proposed method with DIPP \cite{huang2023differentiable}. Notice that DIPP features a structured learning framework that integrates motion prediction and planning, utilizing a pre-configured reference route while excluding decision-making from the framework. By comparing our method with this recent benchmark in the field, we aim to validate the efficacy of our approach and emphasize the importance of incorporating decision-making within an integrated framework.
        \end{enumerate}


        \begin{figure*}[t]
		\centering
		\subfloat[DIPP]{\includegraphics[width=0.25\linewidth]{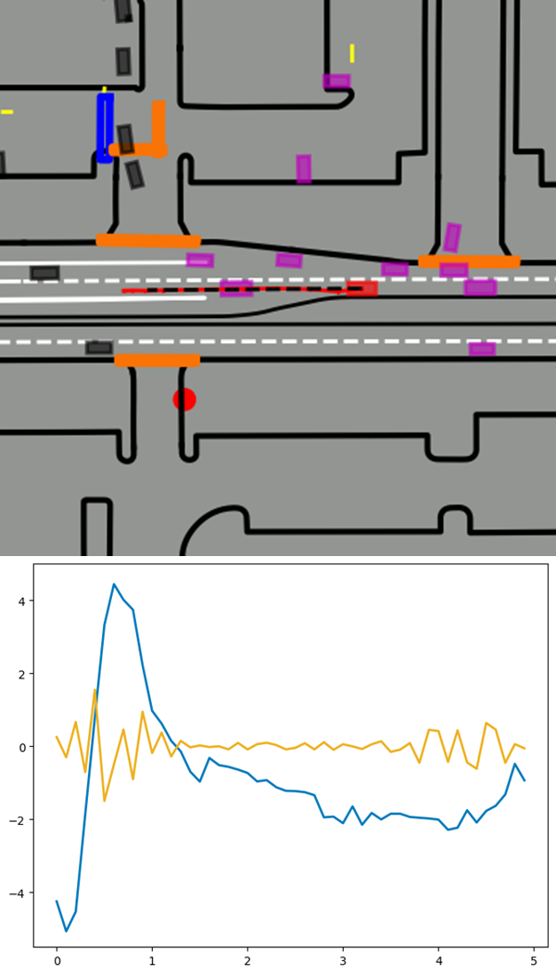}
        \label{fig:1_dipp}
        }
        	\subfloat[Ours]{\includegraphics[width=0.25\linewidth]{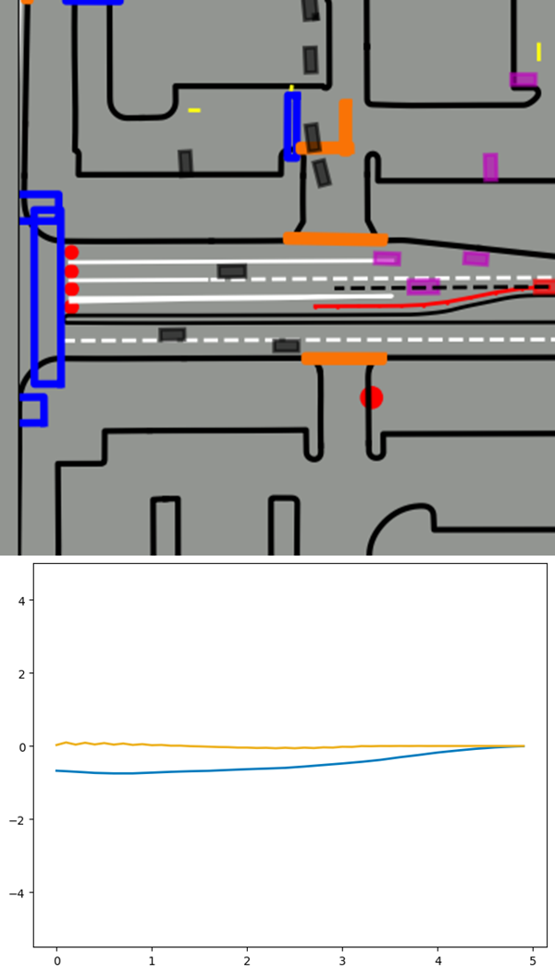}
        \label{fig:1_ours}
        }
        	\subfloat[DIPP]{\includegraphics[width=0.25\linewidth]{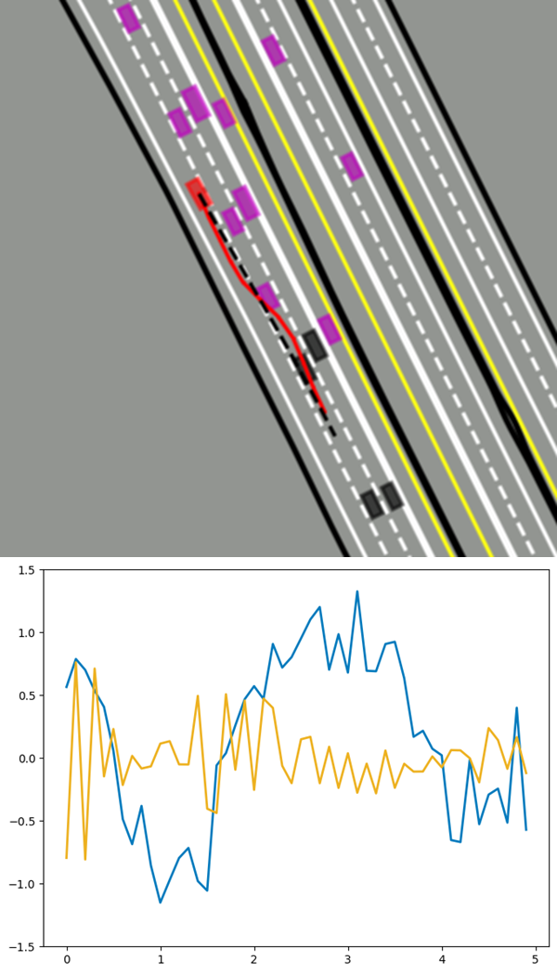}
        \label{fig:2_dipp}
        }
        	\subfloat[Ours]{\includegraphics[width=0.25\linewidth]{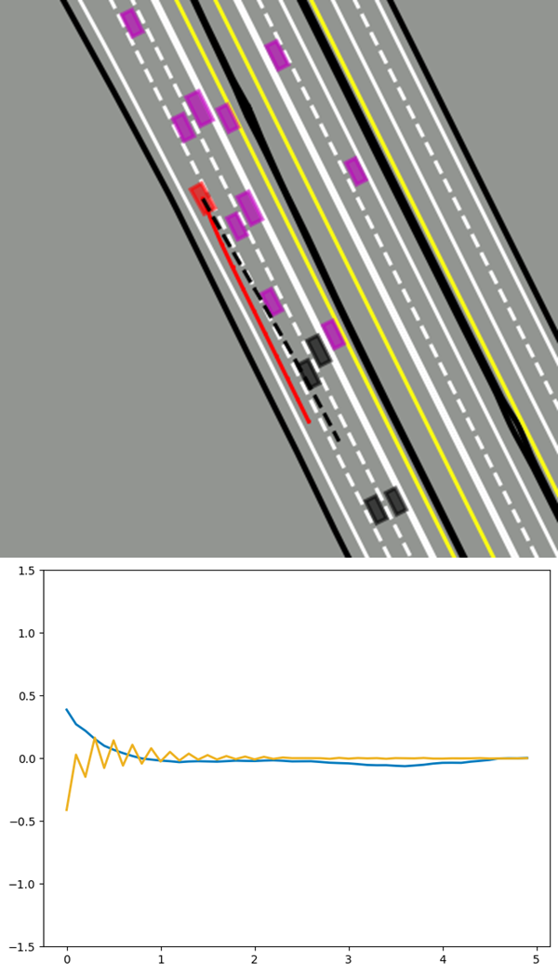}
        \label{fig:2_ours}
        }
		\caption{Comparison of the planning outcomes by DIPP and our proposed framework in the open-loop testing. The top figures show the planned trajectory, with the red solid lines showing the planned trajectories for the AV, and the black dotted lines representing the reference line from expert demonstration. The bottom figures plot the control inputs of the AV ($y$-axis) across the planning horizon in seconds ($x$-axis), with the blue line denoting the acceleration, and the yellow line denoting the steering angle.}
		\label{fig:ol_comparison}
	\end{figure*}

        \begin{table}[t]
          \centering
          \caption{comparison of the proposed method with baseline methods in open-loop testing}
          \resizebox{1.0\linewidth}{!}{
            \begin{tabular}{c|cc|cc|ccc}
            \toprule
            \multirow{2}[2]{*}{Method} & Collision & Off route & Acc.  & Lat. Acc. & \multicolumn{3}{c}{Planning error (m)} \\
                  & ($\%$)  & ($\%$)  & ($m/s^2$) & ($m/s^2$) & @1s   & @3s   & @5s \\
            \midrule
            Vanilla IL & 16.83 & 21.48 & 0.541 & 0.150 & 1.727 & 1.966 & 3.699 \\
            IL + OPT & 8.71  & 6.53  & 0.411 & 0.048 & 0.611 & 3.092 & 6.797 \\
            DIPP  & 7.55  & 7.68  & 0.794 & 0.046 & 0.410 & 1.715 & 4.630 \\
            Ours  & 4.50  & 8.85  & 0.336 & 0.039 & 0.378 & 2.105 & 4.763 \\
            \bottomrule
            \end{tabular}%
          }\label{tab:ol}%
        \end{table}%

      We first consider open-loop testing where the evaluation is based on the ground truth to measure the similarity of the results produced by the models and the trajectories by human drivers.
        The open-loop comparison between our proposed method and the alternative methods are presented in Table~\ref{tab:ol}.

        The collision rate and off-route percentage serve as key indicators of the model's safety performance. Our method achieves the lowest collision rate at 4.50\%. 
        The slightly higher off-route percentage is because the AV in our proposed method is not provided with a reference route, and the AV automatically selects the lane of travel based on driving performance consideration.
        Planning error indicates the accuracy of the planned trajectory relative to human trajectories. Our method performs well at shorter time horizons, with a planning error of 0.378 m at 1 second, comparable to DIPP (0.410 m) and significantly better than Vanilla IL (1.727 m). However, at longer horizons (3 and 5 seconds), the planning error for our method increases, surpassing that of DIPP, while remaining competitive with the Vanilla IL and IL + OPT methods. In terms of comfort, our method also demonstrates strong performance, as evidenced by favorable performance for longitudinal and lateral acceleration, indicating a good balance between comfort and responsiveness.
        
        From the testing results, it appears that our method excels in short-term planning, but is prone to error accumulation in long term. We wish to highlight that this is not a limitation of the proposed method. In contrast, the ability to make optimized decisions is a key feature of our end-to-end planning framework. Our proposed method makes optimized decisions, resulting in planned trajectories that may diverge from human trajectories, yet offer enhanced driving performance.

 \textbf{Comparison with DIPP:}
        To further illustrate the importance of decision-making ability, we compare the distinction between DIPP and our proposed method in Fig.~\ref{fig:ol_comparison}, noting that the decision-making is missing from DIPP and the planning is circumscribed by the reference line.
        In the same scenario with an LV ahead, the planning result by the DIPP method highly mimics the ground truth trajectory (Fig.~\ref{fig:ol_comparison}\subref{fig:1_dipp}), while our proposed method can make decision to change to the right lane (Fig.~\ref{fig:ol_comparison}\subref{fig:1_ours}). The visualization of the control inputs according to the two different planning outcomes are shown at the bottom of Fig.~\ref{fig:ol_comparison}\subref{fig:1_dipp} and Fig.~\ref{fig:ol_comparison}\subref{fig:1_ours}. 
        Both the steering angle and acceleration exhibit larger fluctuations in DIPP compared to our proposed method, indicating that the inclusion of decision-making in our end-to-end planning framework leads to more stable control inputs and enhanced driving performance.
        In a scenario where lane keeping is more appropriate, the proposed method opts to maintain the current lane since the left lane is already occupied (Fig.~\ref{fig:ol_comparison}\subref{fig:2_ours}), rather than rigidly following the reference line that suggests a left lane change (Fig.~\ref{fig:ol_comparison}\subref{fig:2_dipp}).
        Referring to the control inputs by DIPP and our proposed method at the bottom of Fig.~\ref{fig:ol_comparison}\subref{fig:2_dipp} and Fig.~\ref{fig:ol_comparison}\subref{fig:2_ours}, an unnecessary lane change due to lack of decision-making ability in DIPP results in volatile control values and inevitably undermines driving experience. In contrast, our proposed method addresses this decision-making problem, leading to a more desirable driving plan.

        Overall, our method exhibits a robust balance across all evaluated metrics, consistently outperforming both Vanilla IL and IL + OPT methods, while closely matching the performance of DIPP. It achieves commendable safety outcomes, attaining the lowest collision rate. Additionally, our method demonstrates competitive accuracy in short-term planning. The observed trade-off in long-term planning accuracy can be attributed to the optimized decision-making process, which does not hinder performance in closed-loop testing.

    
    In contrast to open-loop testing, closed-loop testing enables AV to execute actions different from those recorded in the human trajectory. This aspect underscores our model's capability to autonomously make discretionary decisions at enhancing safety and traveling efficiency, while delivering riding comfort at the same time. 
    The results of the closed-loop testing are summarized in Table~\ref{tab:cl}, which presents comparisons with baseline and ablation methods. We first discuss the comparative results with the baseline methods, and continue the discussion on the ablation study in the next subsection.

    \begin{table*}[t]
      \centering
      \caption{Comparison of the proposed method with baseline and ablation methods in closed-loop testing}
        \begin{tabular}{c|cc|cc|cc}
        \toprule
        \multicolumn{1}{c|}{\multirow{2}[2]{*}{Method}} & \multicolumn{2}{c|}{(a) Driving Safety} & \multicolumn{2}{c|}{(b) Traveling Efficiency} & \multicolumn{2}{c}{(c) Riding Comfort} \\
              & Collision ($\%$) & Safety Index & Progress (m) & Speed (m/s) & Acc. ($m/s^2$) & Lat. Acc. ($m/s^2$) \\
        \midrule
        Vanilla IL & 22    & 7.28  & 9.89  & 13.02 & 1.215 & 0.639 \\
        IL + OPT & 9     & 8.29  & 26.82 & 3.66  & 0.614 & 0.168 \\
        DIPP  & 5     & 13.48 & 47.58 & 5.11  & 0.705 & 0.092 \\
        \midrule
        No initialization on decision & 5     & 15.37 & 48.91 & 4.55  & 0.786 & 0.038 \\
        No initialization on action & 6     & 17.57 & 56.47 & 5.83  & 0.371 & 0.178 \\
        No learnbale cost function & 9     & 6.23  & 49.09 & 6.48  & 0.507 & 0.168 \\
        No integrated prediction & 9     & 6.32  & 54.74 & 5.75  & 0.635 & 0.187 \\
        \midrule
        Ours  & 4     & 17.42 & 72.77 & 6.22  & 0.591 & 0.046 \\
        \bottomrule
        \end{tabular}%
      \label{tab:cl}%
    \end{table*}%

    In terms of driving safety, the proposed method exhibits the lowest collision rate (4\%) among all methods, with a high safety index averaged at 17.42, outperforming methods such as Vanilla IL (22\%) and the benchmark method DIPP (5\%).
    In terms of progress and speed, indicators of traveling efficiency, the proposed method makes substantial progress (72.77 m), which is the highest among all methods. The longer distance covered is supported with a higher speed (6.22 m/s) in average, as compared to other baselines, such as DIPP (5.11 m/s).
    The riding comfort, measured through longitudinal and lateral accelerations, remains within reasonable comfort levels.
    
    As summary, the proposed method has demonstrated a low collision rate, significant traveling efficiency, and acceptable comfort levels in closed-loop testing, indicating that it outperforms existing baselines. 
    The enhancements observed in the proposed method can be attributed to three key factors:
    1) Joint learning of related tasks: The closely related driving tasks of prediction, decision-making, and trajectory planning are trained jointly, contributing to overall driving performance. The upstream task (i.e., prediction) is informed by the ultimate outcomes, enabling it to generate results that align closely with the desired objectives, thereby facilitating more efficient and goal-oriented travel.
    2) Integrated decision-making: The incorporation of decision-making within the integrated framework allows the AV to autonomously make choices that prioritize safety and traveling efficiency, enabling the AV to optimize driving performance within a data-driven framework.
    3) Optimization constraints: The optimization formulation within the proposed framework introduces critical constraints that enhance safety and ensure the successful completion of driving tasks. The optimizer also corrects the initialized driving plan generated by the prediction module. This capability is particularly beneficial in closed-loop testing, as such corrections help to prevent error accumulation when the AV encounters unforeseen scenarios.

\begin{figure*}[t]
		\centering
		\subfloat[]{\includegraphics[width=0.25\linewidth]{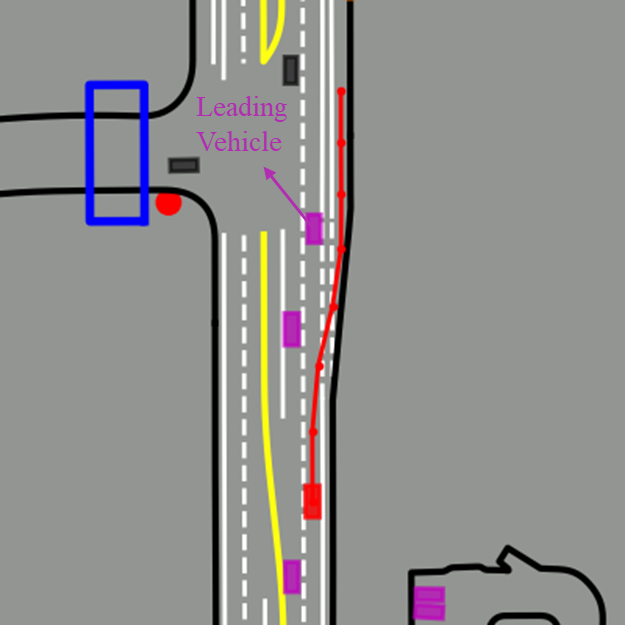}
        \label{fig:scenarios1}
        }
        	\subfloat[]{\includegraphics[width=0.25\linewidth]{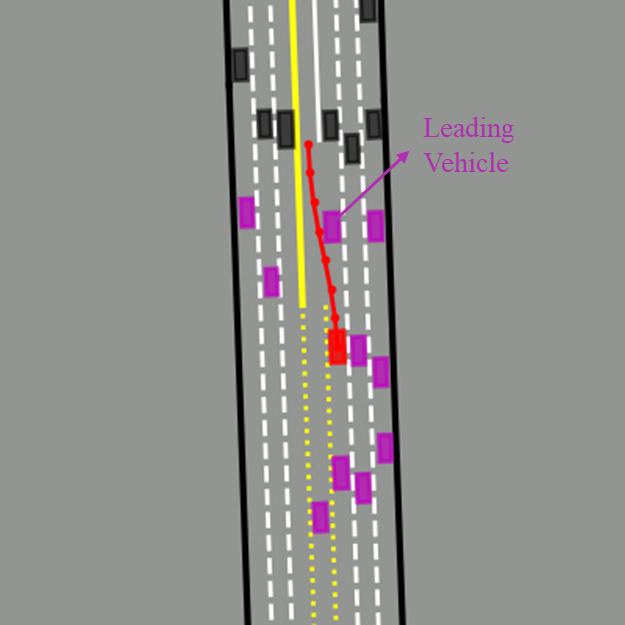}
        \label{fig:scenarios2}
        }
        	\subfloat[]{\includegraphics[width=0.25\linewidth]{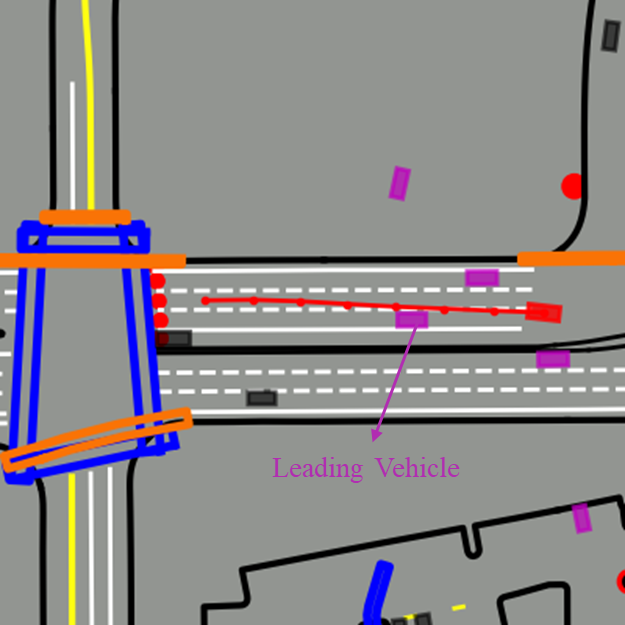}
        \label{fig:scenarios3}
        }
        	\subfloat[]{\includegraphics[width=0.25\linewidth]{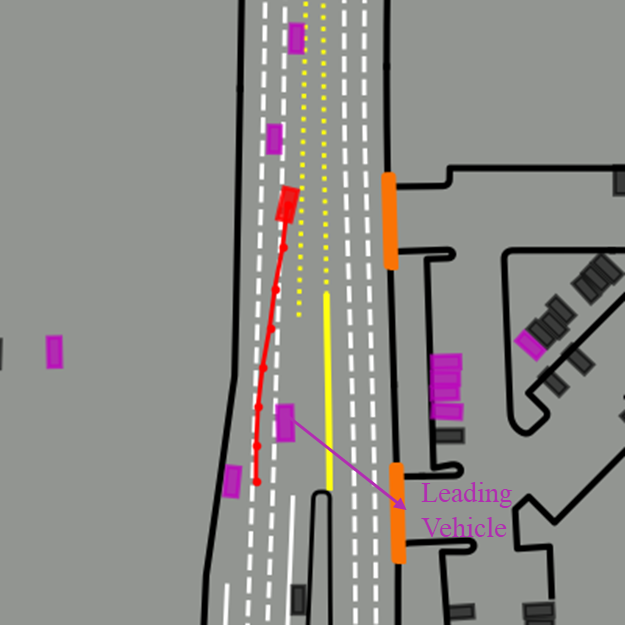}
        \label{fig:scenarios4}
        }
        \vspace{-4mm}
        	\subfloat[]{\includegraphics[width=0.25\linewidth]{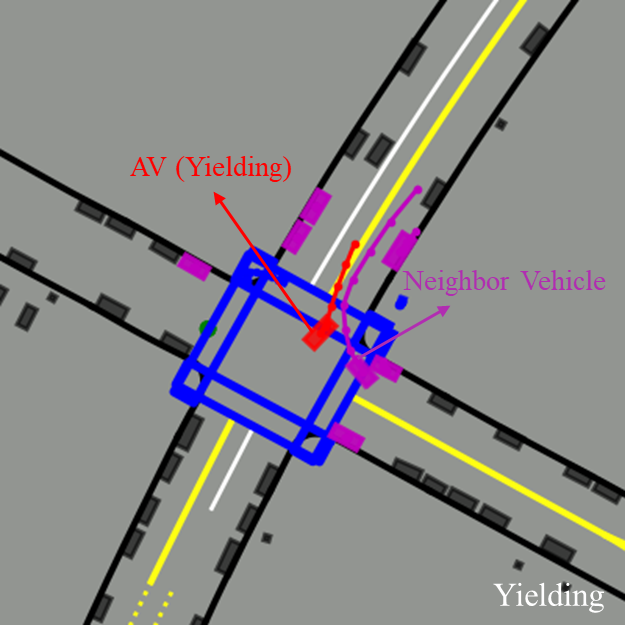}
        \label{fig:scenarios5}
        }
        	\subfloat[]{\includegraphics[width=0.25\linewidth]{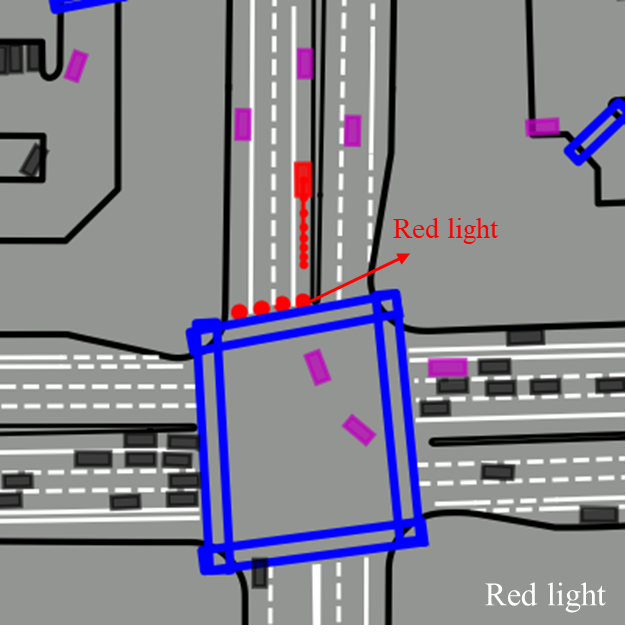}
        \label{fig:scenarios6}
        }
        	\subfloat[]{\includegraphics[width=0.25\linewidth]{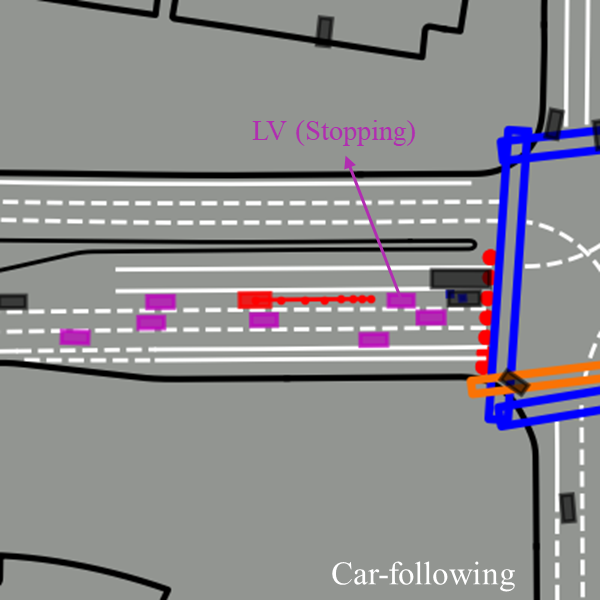}
        \label{fig:scenarios7}
        }
        	\subfloat[]{\includegraphics[width=0.25\linewidth]{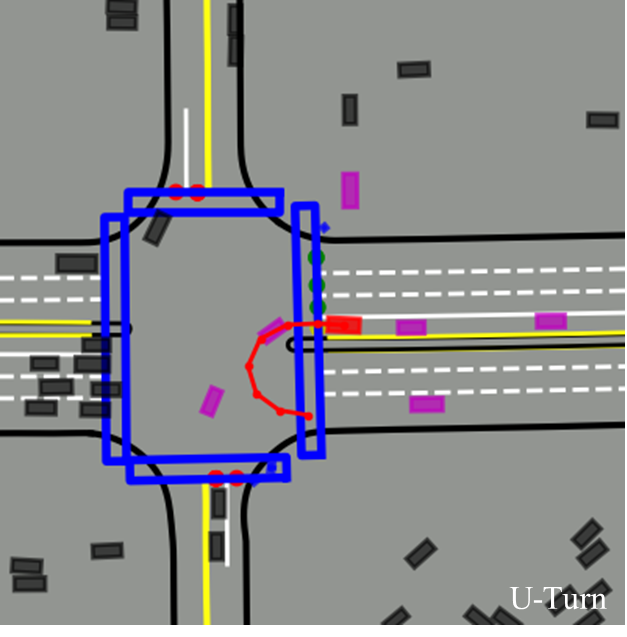}
        \label{fig:scenarios8}
        }
		\caption{Representative scenarios of the proposed framework in closed-loop testing. The red solid lines are the planned trajectories for the AV. Top: optimized lane-changing maneuvers with LV on the same lane. Bottom: typical urban driving scenes include yielding, compliance with traffic light, car-following, and U-turn.}
		\label{fig:scenarios}
	\end{figure*}

    We present representative scenarios from our close-loop testing in Fig.~\ref{fig:scenarios}. 
    Four scenarios in the top panel show that the proposed framework can generate driving plans on optimized lane-changing maneuvers and respective trajectories to navigate the AV toward available lanes with better traveling conditions. 
    The four scenarios described at the bottom demonstrate typical urban driving situations, including yielding to vehicle at intersections, stopping at red traffic signals, executing car-following maneuvers, and performing U-turns.
    These scenarios are critical for evaluating the decision-making capabilities and safety performance of AVs in complex driving environments.

    \subsection{Ablation Study} \label{sec:ablation}
        The ablation study is conducted to validate the efficacy of key components in the proposed framework. Four categories of ablation are conducted to assess the impact of removing specific elements, compared with the baseline, which represents the complete implementation of the proposed method:
        \begin{enumerate}
            \item {No initialization on decision}: The optimizer begins with a lane-keeping decision rather than utilizing decisions from the upstream prediction module.
            \item {No initialization on action}: The initialization of control inputs for the AV by the upstream prediction module is removed by setting these inputs to zero and transforming them to trajectory outputs using the vehicle model.
            \item {No integrated prediction}: The prediction module in the proposed framework is replaced by a simplified assumption that agents travel along their lanes at constant velocities.
            \item {No learnable cost function}: The weighting components in the cost function is manually engineered instead of being learned from training data.
        \end{enumerate}

        We conduct closed-loop testing for ablation studies, with the results included in Table~\ref{tab:cl}.
        The testing results indicate that removing the initialization for decision-making and action does not lead to a significant deterioration in safety. The collision rates for these configurations are reported at 5\% and 6\%, respectively, while the average safety indices are 15.37 and 17.57. Although the collision rates are slightly higher, the safety indices remain comparable to those observed with the full implementation of the proposed method. This performance in safety indices can be attributed to the relatively low speed of AV in both ablation scenarios. The ablations on initialization for decision-making and action demonstrate that these components play a crucial role in traveling efficiency, while a basic level of safety is still achievable without them.

        In contrast, the ablation of the learnable component in the cost function and the integrated prediction shows a more pronounced impact. Each of these ablation methods experiences a collision rate as high as 9\%, resulting in significant declines in both safety indices and traveling efficiency. This underscores the importance of integrated prediction and the learnable cost function as mechanisms that enable the model to better anticipate and adapt to complex driving environments, which is consistent with the findings in~\cite{huang2023differentiable}.


    \subsection{Discussion}
    We provide an analysis centering on the impact of key features of our proposed method on driving performance in this subsection. The features considered include: 1) Optimized decision-making, 2) Learning of initialized decisions, and 3) Management of hard constraints within the learning framework. In closed-loop testing, safety is quantified by the completion rate, which indicates the extent to which AV successfully navigates the testing scenario without collisions, boundary violations, or failures resulting from an inability to find a solution to the optimization problem. Traveling efficiency is measured by the progress made in meters. Additionally, we report the convergence of the optimizer by determining whether it meets the predefined threshold for termination within the maximum number of iterations. This metric reflects the optimizer's performance in solving the optimization problem and does not necessarily indicate a failure of the driving task.

    \subsubsection{Effects of Optimized Decisions} 
    The integration of optimized decision-making is a crucial aspect of the proposed framework. We analyze the impact of optimized lane-changing decisions on driving performance, with results detailed in Table~\ref{tab:lc}. The metric Optimized Lane Change Rate (OLC) (\%) refers to the lane changes induced by the optimized decision-making process. Notably, the DIPP method does not incorporate this optimized decision-making process and the lane changes in DIPP are learned from demonstrations, so the OLC rate in DIPP is not taken into consideration.
    In contrast, our framework facilitates an optimized lane-changing maneuver rate of up to \(57\%\). This enhancement is driven by our method's ability to enable the AV to make decisions aligned with the specified performance objectives. Our experiment results demonstrate improved safety, as evidenced by higher completion rates, and greater traveling efficiency, reflected in increased progress. Additionally, our proposed method maintains a convergence rate comparable to that of the DIPP method, despite utilizing a more complex optimization formulation.

    \begin{figure}[t]
        \centering
         \subfloat[Ours]{\includegraphics[width=0.5\linewidth]{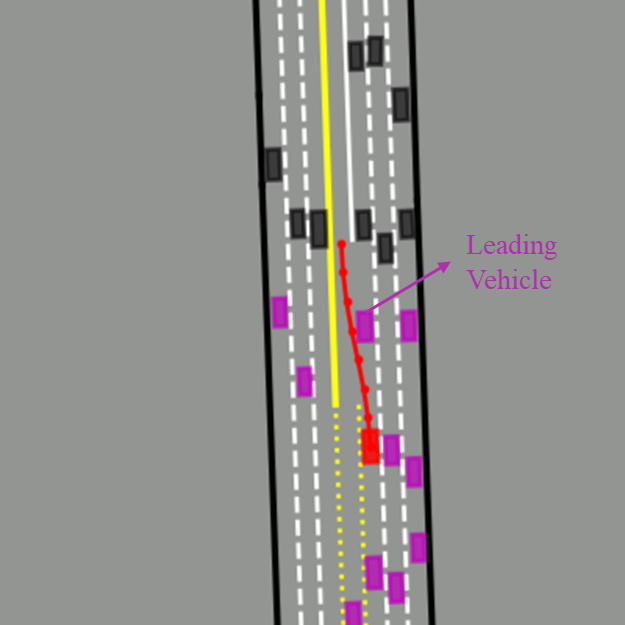}
        \label{fig:ol_noDecision1}
        }
        \subfloat[No initialized decision]{\includegraphics[width=0.5\linewidth]{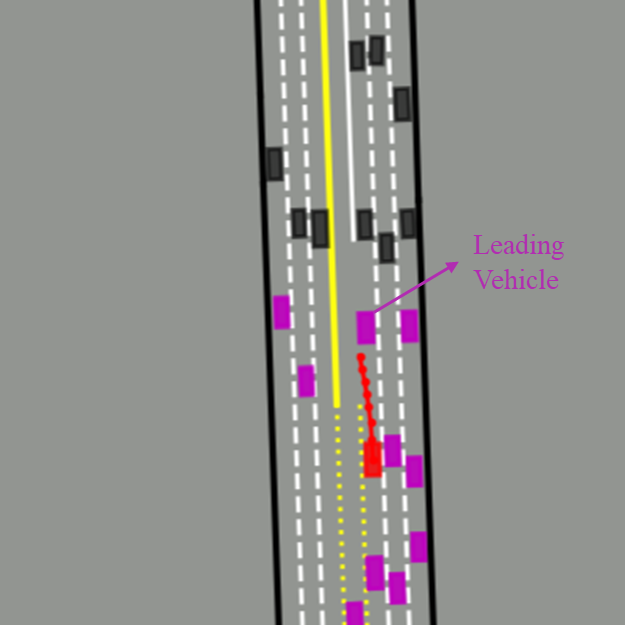}
        \label{fig:ol_noDecision2}
        }
        \caption{Comparison of the planned trajectory with and without the learned initialization of decisions in closed-loop testing.}
        \label{fig:cl_noDecision}
    \end{figure}

        \begin{table}[t]
          \centering
          \caption{Effects of optimized lane change decisions to the driving performance}
          \resizebox{1.0\linewidth}{!}{
            \begin{tabular}{c|cccc}
            \toprule
            Method & OLC (\%) & Complete (\%) & Converge (\%) & Progress (m) \\
            \midrule
            DIPP  & -     & 87    & 100   & 47.58 \\
            No initialization on decision & 84    & 92    & 15    & 48.91 \\
            Ours  & 57    & 94    & 100   & 72.77 \\
            \bottomrule
            \end{tabular}%
          }\label{tab:lc}%
        \end{table}%
        
    \subsubsection{Learning of Initialized Decisions}
        The initialization of decisions from the prediction module is a learnable component of the proposed method. Our analysis reveals that the learning of these initialized decisions is critical to the optimization problem's performance and contributes to the overall driving performance. We compare the driving performance of the fully implemented method against one without decision initialization in Table~\ref{tab:lc}. The absence of this initialization results in a higher rate of lane changes, yet it negatively affects the completion rate and overall progress. This suggests that the lane changes in the ablation study without decision initialization are unnecessary and compromise overall driving performance. Furthermore, the performance of the optimization problem significantly declines, dropping to \(15\%\). 
        As illustrated in Fig.~\ref{fig:cl_noDecision}, a comparative analysis is conducted through closed-loop testing, which substantiates that the proposed method effectively incorporates the LV into the decision-making process, thereby generating a safe and viable trajectory. In contrast, the planning outcome in the absence of an initialized decision is markedly more conservative, underscoring the adverse impact on the optimizer's performance when initialization is not provided.       
        These findings highlight the importance of learning initialized decisions in ensuring the reliable convergence of the optimizer, thereby facilitating safe and efficient driving performance.

    \subsubsection{Decision-Making Constraint Compliance}
    The decision-making problem is bound by the integer and equality constraints as outlined by~\eqref{eq:binary} and~\eqref{eq:sumb_1}. The weights used to enforce these constraints are set to $w_{bi} = 10$ and $w_{eq} = 1000$ in our method, respectively. We conduct comparative analyses by setting the associated weights to $w_{bi} = 1$ and $w_{eq} = 100$, respectively. The testing results are summarized in Table~\ref{tab:hardC}.
    Our chosen weights ensure a compliance rate of 100\% with the constraints in the decision-making problem. In contrast, reducing the weights significantly deteriorates the compliance rate. Adherence to these hard constraints is critical for the ultimate driving performance; violations are associated with decreases in both completion rates and overall progress.

    \begin{table}[t]
      \centering
      \caption{Compliance on decision-making constraint and impacts on driving performance}
      \scriptsize
        \begin{tabular}{cccc}
        \toprule
              & Compliance (\%) & Complete (\%) & Progress (m) \\
        \midrule
        $w_{bi} = 1, w_{eq} = 1000$ & 86    & 70    & 60.98 \\
        $w_{bi} = 10, w_{eq} = 100$ & 88    & 68    & 59.42 \\
        Ours  & 100   & 83    & 72.77 \\
        \bottomrule
        \end{tabular}%
      \label{tab:hardC}%
    \end{table}%

\section{Conclusion} \label{sec:con}
    We propose an end-to-end planning framework that focuses on motion prediction, decision-making, and trajectory planning tasks in AD. 
    Our hybrid approach leverages the adaptability of the deep learning approach while simultaneously employing optimization techniques to ensure that the learning process is guided by predefined optimizing objectives.
    By fostering this synergy, the proposed AD system can attain effective performance in diverse scenarios by learning from expert demonstrations.
    Our experimental results show that the proposed method excels in pursuing performance in terms of safety, traveling efficiency, and riding comfort that go beyond these demonstrations. Overall, by integrating the decision-making capability into the end-to-end planning system, the proposed method enhances the driving experience and paves the way for more robust and reliable AD solutions.
    One promising future direction involves incorporating hard constraints within the structured learning framework, by employing optimization techniques such as the augmented Lagrangian method. 
    Additionally, it is important to address the lengthy computation time associated with differentiable optimizers. Research on developing more efficient solvers to enhance the performance and applicability of differentiable optimization in this context is also one of the possible future directions.

\bibliographystyle{ieeetr}
\bibliography{mybibfile}

\end{document}